# Logical Assessment Formula and Its Principles for Evaluations with Inaccurate Ground-Truth Labels

Yongquan Yang

Email address: remy_yang@foxmail.com
Affiliation: Yang Family, Chengdu, China

**Abstract**

Evaluations with accurate ground-truth labels (AGTLs) have been widely employed to assess predictive models for artificial intelligence applications. However, in some specific fields, such as medical histopathology whole slide image analysis, it is quite usual the situation that AGTLs are difficult to be precisely defined or even do not exist. To alleviate this situation, we propose logical assessment formula (LAF) and reveal its principles for evaluations with inaccurate ground-truth labels (IAGTLs) via logical reasoning under uncertainty. From the revealed principles of LAF, we summarize the practicability of LAF: 1) LAF can be applied for evaluations with IAGTLs on a more difficult task, able to act like usual strategies for evaluations with AGTLs reasonably; 2) LAF can be applied for evaluations with IAGTLs from the logical perspective on an easier task, unable to act like usual strategies for evaluations with AGTLs confidently.

Key words: Evaluation of predictive models; Evaluation with inaccurate ground-truth labels; Reasoning under uncertainty

## 1. Introduction

Assessment of predictive models is an important foundation for investigating potentially appropriate solutions for various artificial intelligence applications. The most usual strategy is to evaluate predictive models based on massive testing samples with accurate ground-truth labels (AGTLs), i.e., the strategy of evaluations with massive AGTLs [1–3]. Due to the difficulties of collecting massive testing samples with AGTLs, the strategy of evaluations with limited AGTLs [4, 5] has also been proposed. Both strategies require more or less AGTLs to be carried out. However, in some specific fields, such as medical histopathology whole slide image analysis (MHWSIA), it is quite usual the situation that AGTLs are difficult to be precisely defined [6, 7] or even do not exist [8–10], which makes the collection of AGTLS for evaluations of unseen testing results much more difficult. This situation constraints the usage of the strategies of evaluations with massive AGTLs or limited AGTLs.

To alleviate this, researchers have begun to explore the usage of inaccurate ground-truth labels (IAGTLs) to assess predictive models, i.e., the strategy of evaluations with IAGTLs. Two feasible types of methods have emerged for the strategy of evaluations with IAGTLs. One type of method is to firstly select some probably true targets from the inaccurate targets [11] within the given IAGTLs via probabilistic estimation, and then to achieve evaluations of unseen testing results by referring to the selected probably true targets [12, 13]. The other type of method is to achieve evaluations of unseen testing results by referring to the inaccurate targets [14] within the given IAGTLs with provided or estimated minimal rate of error corresponding to the underlying true targets. Fundamentally, the assumption for these two types of methods is that there are true targets exist in the inaccurate targets represented by the given IAGTLs. In other words, the two types of methods assume that the underlying true targets can be precisely defined, otherwise no probably true targets can be selected from the inaccurate targets within the given IAGTLs and no minimal rate of error corresponding to the underlying true targets can be provided or estimated from the inaccurate targets within the given IAGTLs. Intrinsically, this fundamental assumption makes these two types of methods not suitable for the situation where the underlying true targets are difficult to be precisely defined or even do not exist, such as some applications in the field of MHWSIA [15, 16].

With notice of this, our motivation in this article is to propose a new method for evaluations with IAGTLs which does not need the prerequisites that the existing two types of methods need. That is, the proposed method can achieve evaluations directly with IAGTLs, without the fundamental assumption that the underlying true targets can be precisely defined. We propose logical assessment formula (LAF) that evaluates the predicted targets by referring to multiple inaccurate targets, which are extracted from the given IAGTLs. The multiple inaccurate targets extracted from the given IAGTLs are required to merely contain various information consistent with our prior knowledge about the underlying true targets, which can be conveniently provided by experts when preparing IAGTLs for a specific task. As a result, the proposed LAF is different from the existing two types of methods for evaluations with IAGTLs, and more detailed comparisons between LAF and existing two types of methods for evaluations with IAGTLs are shown as Table. 1.

Table 1. Comparisons between LAF and existing two types of methods for evaluations with IAGTLs. TT is short for true target.

| Method | Assumption for TT | Preparation for evaluation | Evaluation procedure |
| --- | --- | --- | --- |
| Type one | TTs exist in IAGTLs | Selecting some probable TTs from IAGTLs | Evaluating on the probable TTs selected from IAGTLs |
| Type two | TTs exist in IAGTLs | Providing / estimating rate of TT error in IAGTLs | Evaluating on IAGTLs regarding to the provided / estimated rate of TT error |
| LAF | None | Extracting multiple inaccurate targets from IAGTLs | Evaluating on the multiple inaccurate targets extracted from IAGTLs |

Specifically, LAF constitutes of three processing procedures, including logical fact narration, logical consistency estimation and logical assessment metric build. The former two processing procedures of LAF were inspired from the prerequisites for the abduction of abductive learning [17]. The logical fact narration procedure narrates a list of logical facts about the underlying true targets from the multiple inaccurate targets. The logical consistency estimation procedure estimates a list of logical consistencies between the predicted targets and the logical facts about the underlying true targets narrated by the logical fact narration procedure. The logical assessment metric build procedure derives a series of abstractly formalized metrics from the logical consistencies estimated by the logical consistency estimation procedure to represent the evaluations of the predicted targets compared with the underlying true targets.

We have also revealed the principles of LAF for evaluations with IAGTLs via logical reasoning under uncertainty, which is a series of logical processes with reasoning under uncertainty [18–22]. In addition, from the revealed principles of LAF, we have summarized the practicability of LAF for the situation, where AGTLs are not available while multiple inaccurate targets containing various information consistent with our prior knowledge about the underlying true targets can be extracted from the available IAGTLs. The summarized practicability of LAF include: 1) LAF can be applied for evaluations with IAGTLs on a more difficult task, able to act like usual strategies for evaluations with AGTLs reasonably; and 2) LAF can be applied for evaluations with IAGTLs simply from the logical point of view on an easier task, unable to act like usual strategies for evaluations with AGTLs confidently.

The rest contents are structured as follows. In Section 2, we review related works. In Section 3, we formulate the methodology of LAF and related extensions. In Section 4, we reveal the principles of LAF for evaluations with IAGTLs. In Section 5, we summarize the practicability of LAF from the revealed principles. In Section 6, we discuss the whole article. Finally, we point out some works related to LAF that should be done in the future in Section 7.

## 2. Related Work

## 2.1. Evaluations with massive AGTLs

To implement the strategy of evaluations with accurate ground-truth labels (AGTLs), some evaluation metrics need be defined based on the predictions of the provided testing samples and corresponding AGTLs. For instance, referring to the similarity and difference between the predictions of the testing samples and corresponding AGTLs, basic evaluation metrics such as true positive (TP), false positive (FP), true negative (TN) and false negative (FN) can be defined. Employing these basic evaluation metrics, more evaluation metrics such as precision, recall, f-measure, accuracy and intersection of union can be further defined for the case of evaluations for classification [3], image segmentation [1, 2] and etc.

## 2.2. Evaluations with limited AGTLs

To implement the strategy of evaluations with limited AGTLs, Jung et al., [2] proposed an approach based on a set of pre-trained classifiers, a ground-truth label set corresponding to the testing examples, and multiple extra pseudo ground truth label sets, which can possibly be like Kandinksy Patterns [23]. The predictions of the testing samples via the set of pre-trained classifiers are ranked on the corresponding ground-truth label set, and are also respectively ranked on the multiple extra pseudo ground truth label sets. The rankings on the multiple extra pseudo ground truth label sets are respectively compared with the ranking on the ground-truth label set by measurements of correlation and coefficient. The pseudo ground truth label set corresponding to the best measurement of correlation and coefficient can be selected for evaluation without expert labels. Deng et al., [5] proposed to regress the accuracy of a classifier on a testing dataset to achieve evaluations with LAGTLs. Firstly, a classifier $f(\theta)$ is trained with a given data set $x$ and its corresponding label set $y$. Then, given a testing data set $x_t$ and corresponding accurate ground-truth label set $y_t$, the performances of $f(\theta)$ on $x_t$ were obtained via evaluations metrics defined on AGTLs. Secondly, multiple testing data sets and corresponding label sets were generated via data augmentation technologies, and a number of performances of $f(\theta)$ on the generated multiple testing data sets can be calculated. Finally, using the feature statistics to represent the generated testing data sets, a regression model was trained for prediction of the model performance.

## 2.3. Evaluations with IAGTLs

To implement the strategy of evaluations with inaccurate ground-truth labels (IAGTLs), Warfield et al., [11] proposed simultaneous truth and performance level estimation (STAPLE). Firstly, based on expectation-maximization (EM) algorithm [24] and related works [25], STAPLE estimates the probabilities of being true targets for the inaccurate targets within the given IATGLs, which may be the results of a human rater or the predictions of an appropriately trained predictive model. Then, the probably true targets estimated by STAPLE can be further leveraged for evaluations of unseen testing results [12, 13] without requiring AGTLs. Joyce et al., [14] proposed approximate ground truth refinement (AGTR), which is an incomplete clustering of a dataset for probable ground truth labels. More specifically, data points belonging to the same cluster share the same underlying ground truth labels. To create an AGTR, one requires domain knowledge effort to group "alike" data points with high fidelity and must accounts for a minimal rate of error inherent to the grouping. Because the fact that a minimal rate of error corresponding to AGTR must be provided, bounds on specific metrics can be estimated to evaluate the testing results of any system that groups samples into sub-populations. These existing methods for evaluations with IAGTLs are based on the assumption that some true targets probably exist in the inaccurate targets represented by the given IAGTLs. Different from these methods, the LAF proposed in this article does not need such fundamental assumption to achieve evaluations with IAGTLs.

## 2.4. Reasoning under uncertainty

According to Judea Pearl [18], reasoning under uncertainty, which almost can be identified in any field in artificial intelligence, are processes leading from evidences or clues to guesses and conclusions under conditions of partial information. Firstly, he classified the AI researchers tackling the problems of reasoning under uncertainty into three schools: logicist, neo-calculist and neo-probabilist, which captured the computational differences among various approaches existing at the time for reasoning under uncertainty [18]. Then, additionally taking into consideration the semantical properties of various approaches for reasoning under uncertainty, he proposed a more fundamental taxonomy, namely the extensional vs intentional approaches [18]. With this proposed taxonomy, he officially presented a definition of reasoning under uncertainty for computational intelligence. Krause et al [19] proposed a uniform framework tackling reasoning under uncertainty for computational intelligence. Incorporating a number of numerical and symbolic techniques, the proposed framework can assign subjective confidences to propositions referring to their supporting arguments. Developing the qualitative versions of probability theory, possibility theory, and the Dempster-Shafer theory of evidence [26], Parsons [20] provided the means to create qualitative methods for reasoning under uncertainty, which can work with abstractions such as interval values and information about how values change. Based on possibility theory, Dubois et al [21] proposed generalized possibilistic logic (GPL) for epistemic reasoning. Via modelling comparative uncertainty and ignorance, GPL is particularly suitable for reasoning about what an agent knows about the beliefs of another agent [21]. More recently, Ristic et al [22] proposed a framework for the performance assessment of approaches to reasoning under uncertainty, based on the assumption that the approach under investigation is uncertain only due to stochastic variability. Different from these articles that aimed to establish the computational formalisms for reasoning under uncertainty, this article more fundamentally / theoretically / cognitively presents a series of logical processes with reasoning under uncertainty, which aim to reveal the principles of LAF (i.e., logical conclusions that can reflect the practicability of LAF) for evaluations with IAGTLs.

# 3. Logical Assessment Formula

Logical assessment formula (LAF) is presented to achieve reasonable evaluations of the predicted targets for the underlying true targets, which are difficult to be precisely defined, with inaccurate ground-truth labels (IAGTLs).

## 3.1. Definition for Methodology of LAF

The outline for the methodology of LAF is shown as Fig. 1. Details of the formalized definition for the methodology of LAF are presented in the following contents of this subsection.

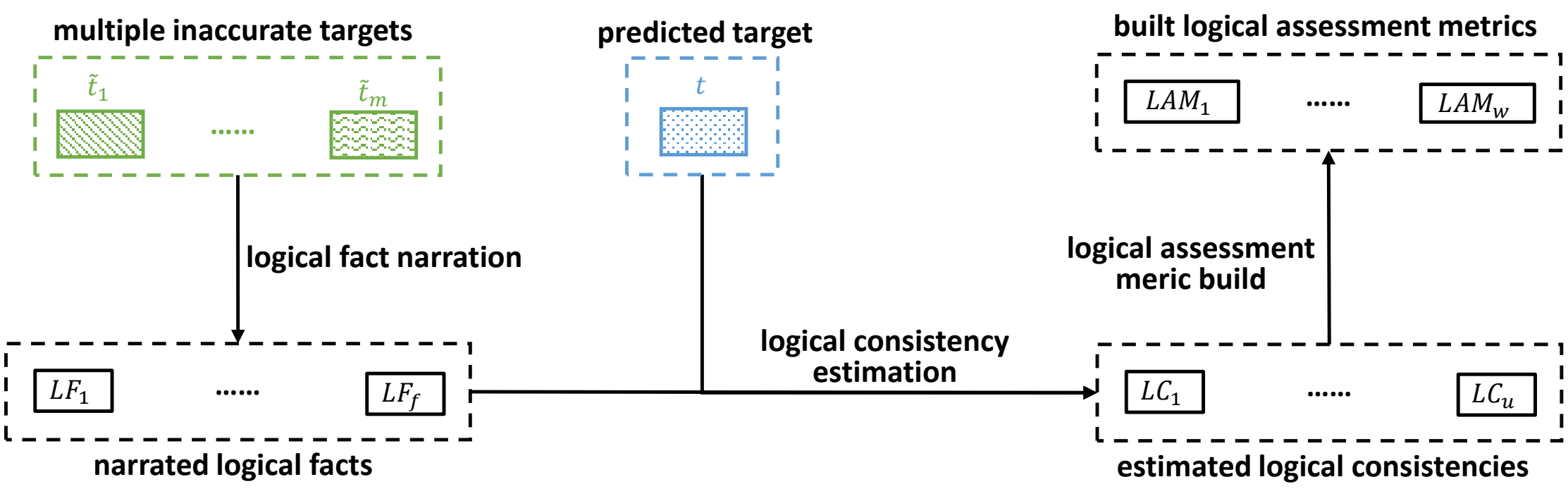

Figure. 1. The outline for the methodology of LAF.

### *3.1.1 Predefined inputs and output of LAF*

As shown in Fig. 1, the inputs of LAF include the predicted target ($t$) for the underlying true target which are difficult to be precisely defined and multiple inaccurate targets ($\tilde{t} = \{\tilde{t}_1, \cdots, \tilde{t}_m\}$) that contain various information consistent with our prior knowledge about the underlying true targets. The inputs $t$ and $\tilde{t}$ are both corresponding to the same instance. As shown in Fig. 1, the output of LAF is a series of logical assessment metrics ($LAM$) that can reflect the evaluations of the predicted target ($t$) compared with the underlying true target.

### *3.1.2 Processing components of LAF*

Based on the predefined inputs and output of LAF, we endow LAF with the aim to produce a series of logical assessment metrics ($LAM$) that reflect the evaluations of predicted target ($t$) compared with the underlying true target by simply referring to multiple inaccurate targets ($\tilde{t}$) that contain various information consistent with our prior knowledge about the underlying true target. To achieve this aim, as shown in Fig. 1, LAF constitutes of three processing components ($PC$), including logical fact narration, logical consistency estimation and logical assessment metric build.

Narrating logical facts ($LF$) from the input multiple inaccurate targets ($\tilde{t}$), the logical fact narration component can be formally expressed as

$$LF = LogicalFactNarrate(\tilde{t}; p^{LFN}) = \{LF_1, \cdots, LF_f\}. \tag{1}$$

Estimating the logical consistencies ($LC$) between the input predicted target ($t$) and the narrated logical facts ($LF$), the logical consistency estimation component can be formally expressed as

$$\begin{aligned} LC &= LogicalConsistencyEstimate(t, LF; p^{LCE}) \\ &= \{LC_1, \cdots, LC_u\}. \end{aligned} \tag{2}$$

Producing a series of logical assessment metrics ($LAM$) based on the estimated logical consistencies ($LC$) between the input predicted target ($t$) and the narrated logical facts ($LF$), the logical assessment metric build component can be formally expressed as

$$\begin{aligned} LAM &= LogicalAssessmentMetricBuild(LC; p^{LAM}) \\ &= \{LAM_1, \cdots, LAM_w\}. \end{aligned} \tag{3}$$

Each $p^*$ of expressions (1)-(3) denotes the hyper-parameters corresponding to the implementation of respective expression.

Note, the narrated $LF$ is a list of qualitative descriptions that logically represent the facts contained in the given multiple inaccurate targets ($\tilde{t}$), the estimated $LC$ is also a list of qualitative descriptions that logically represent the consistencies between the given predicted target ($t$) and the narrated $LF$, and the built $LAM$ is a series of abstractly formalized metrics that are derived from the qualitative descriptions of the estimated $LC$ to represent the evaluations of the predicted target ($t$) compared with the underlying true target.

#### *3.1.3 Formation of LAF*

In summary, the formation of LAF can be formally denoted as

$$LAF \begin{cases} inputs: \begin{cases} t \\ \tilde{t} = \{\tilde{t}_1, \cdots, \tilde{t}_m\} \end{cases} \\ PC \begin{cases} LF = LogicalFactNarrate(\tilde{t}; p^{LFN}) \\ LC = LogicalConsistencyEstimate(t, LF; p^{LCE}) \\ LAM = LogicalAssessmentMetricBuild(LC; p^{LAM}) \end{cases} \\ output: LAM = \{LAM_1, \cdots, LAM_w\} \end{cases}. \tag{4}$$

#### *3.1.4 Usage of LAF*

The suage of LAF can be described as: given the predicted target ($t$) for the underlying true targets which are difficult to be precisely defined and multiple inaccurate targets ($\tilde{t}$) that contain various information consistent with our prior knowledge about the underlying true target, we can obtain, via the processing components of LAF ($LAF\colon PC$), a series of logical assessment metrics ($LAM$) for evaluations of the given predicted target ($t$) compared with the underlying true target. Formally, the suage of LAF can be denoted as

$$\begin{aligned} LAM &= LAF\colon PC(t, \tilde{t}; \{p^{LFN}, p^{LCE}, p^{LAM}\}) \\ &= \{LAM_1, \cdots, LAM_w\}. \end{aligned} \tag{5}$$

## 3.2. Task-specific LAF

Let us consider LAF in the situation, where a specific task is given. Referring to formula (5), the usage of LAF for the task-specific (ts) situation can be formally denoted as

$$\begin{aligned} LAM_{ts} &= LAF\colon PC(t_{ts}, \tilde{t}_{ts}; \{p_{ts}^{LFN}, p_{ts}^{LCE}, p_{ts}^{LAM}\}) \\ &= \{LAM_{ts,1}, \cdots, LAM_{ts,w}\}. \end{aligned} \tag{6}$$

Here, $t_{ts}$ denotes the predicted target for the underlying true targets corresponding to the given specific task, $\tilde{t}_{ts}$ denotes the multiple inaccurate targets that contain various information consistent with our prior knowledge about the underlying true target corresponding to the given specific task, $p_{ts}^*$ denotes the hyper-parameters corresponding to the implementation of respective processing component in formula (4) for task-specific LAF, and $LAM_{ts}$ denotes the constructed logical assessment metrics corresponding to the given specific task. Note, each $p_{ts}^{*}$ in formula (6) needs to be clearly defined for the implementation of respective processing component in formula (4) for the task-specific LAF, according to the inherent characteristics of the given specific task. The output $LAM_{ts}$ here in formula (6) is a series of abstractly formalized metrics corresponding to the given specific task and can be transformed into quantitative evaluations when the input $t_{ts}$ here in formula (6) is designated with a specific method.

### 3.3. Method-specific LAF

Let us consider LAF in the situation, where a specific method is given and the given specific method is particularly for addressing a specific task. On the basis of the task-specific LAF, we can produce a series of method-specific (ms) logical assessment metrics corresponding to the given specific method for a specific task. Referring to formula. (6), the usage of LAF for the method-specific situation can be formally denoted as

$$\begin{aligned} LAM_{ts,ms} &= LAF\colon PC\left(t_{ts,ms}, \tilde{t}_{ts}; \{p_{ts}^{LFN}, p_{ts}^{LCE}, p_{ts}^{LAM}\}\right) \\ &= \left\{LAM_{ts,ms,1}, \cdots, LAM_{ts,ms,w}\right\}. \end{aligned} \tag{7}$$

Here, $t_{ts,ms}$ denotes the predicted targets corresponding to the given specific method for a specific task, and $LAM_{ts,ms}$ denotes the logical assessment metrics corresponding to the given specific method for a specific task. Note, since the input $t_{ts,ms}$ here in formula (7) is designated with a specific method, the output $LAM_{ts,ms}$ here in formula (7) is a series of quantitative evaluations.

### 3.4. LAF-based method performance evaluation

Based on the method-specific LAF, we further establish a LAF-based method performance evaluation (LAF-MPE) strategy. The input of LAF-MPE is a series of logical assessment metrics ($LAM_{ts,ms}$) that are produced by method-specific LAF. The output of LAF-MPE is some method performances ($LMP_{ts,ms}$), which are respectively quantized in range [0,1], to reflect the superiorities of the given specific method for addressing a specific task. Based on the defined input and output of LAF-MEP, we endow LAF-MPE with the aim to derive some logical method performances ($LMP_{ts,ms}$) from a series of logical assessment metrics ($LAM_{ts,ms}$) produced by the method-specific LAF. The processing component ($PC$) of LAF-MPE can be formally expressed as

$$\begin{aligned} LMP_{ts,ms} &= LogicalMethodPerfEval\left(LAM_{ts,ms}; p^{LMPE}\right) \\ &= \left\{LMP_{ts,ms,1}, \cdots, LMP_{ts,ms,v}\right\}, Val(LMP_{ts,ms,v}) \in [0,1]. \end{aligned} \tag{8}$$

Here, $p^{LMPE}$ denotes the hyper-parameters for implementation of formula (8) and $Val(*)$ denotes the value of $*$.

## 4. Principles of LAF for Evaluations with IAGTLs

We reveal the principles of LAF for evaluations with inaccurate ground-truth labels (IAGTLs) via logical reasoning under uncertainty. The revealed principles including the principle of LAF for reflecting logical rationality and the principles of LAF-MPE for reflecting method performance.

### 4.1. Principle of LAF for reflecting logical rationality

We introduce a result (Theorem 1) derived from LAF and reveal the principle of LAF for reflecting logical rationality by proving Theorem 1 with logical reasoning (Proof-T1). Due to limited pages, the proof of Theorem 1 is comprehensively provided in Supplementary.

**Theorem 1**. If the input $t$ of LAF is the prediction for the underlying true target which is indefinable and the input $\tilde{t}$ of LAF has multiple inaccurate targets containing various information consistent with our prior knowledge about the underlying true target, then the output of LAF ($LAM$) can reflect the logical rationality of the input $t$ of LAF compared with the true target.

### 4.2. Principles of LAF-MPE for reflecting method performance

We introduce three results (Theorem 2, Theorem3 and Theorem 4) derived from LAF-MPE and reveal the principles of LAF-MPE for reflecting method performance evaluation by proving Theorem 2, Theorem 3 and Theorem 4 with logical reasoning (Proof-T2, Proof-T3 and Proof-T4). Due to limited pages, the proofs of Theorem 2-4 are comprehensively provided in Supplementary.

**Theorem 2**. If the input $LAM_{ts,ms}$ of LAF-MPE is a series of logical assessment metrics produced by task-specific and method-specific LAF, then the output $LMP_{ts,ms}$ of LAF-MPE can reflect the method performance from the logical point of view.

**Theorem 3**. If the output $LMP_{ts,ms}$ of LAF-MPE can reflect the method performance from the logical point of view and the value of $LMP_{ts,ms}$ is small enough, then the output $LMP_{ts,ms}$ of LAF-MPE can reflect the true method performance reasonably.

**Theorem 4**. If the output $LMP_{ts,ms}$ of LAF-MPE can reflect the method performance from the logical point of view and the value of $LMP_{ts,ms}$ is large enough, then the output $LMP_{ts,ms}$ of LAF-MPE cannot reflect the true method performance confidently.

## 5. Practicability of LAF

Referring to the revealed principles of LAF, we derive a list of practical rules of applying LAF for evaluations with IAGTL on real world tasks. Based on Theorem 1, we have:

**Rule 1**. Referring to multiple inaccurate targets containing various information consistent with our prior knowledge about the true target, LAF can reflect the logical rationality of the prediction for the true target with inaccurate ground-truth labels.

Based on Theorem 2, Theorem 3 and Theorem 4, we have:

**Rule 2**. LAF-MPE can be leveraged to show the method superiority from the logical point of view by computing corresponding $LMP_{ts,ms}$ value.

**Rule 3**. When the computed $LMP_{ts,ms}$ value is small enough, the $LMP_{ts,ms}$ value can reasonably show the true method superiority.

**Rule 4**. When the computed $LMP_{ts,ms}$ value is large enough, the $LMP_{ts,ms}$ value cannot confidently show the true method superiority.

Rule 1 indicates that LAF is suitable for evaluations of the situation where accurate ground-truth labels are not available while multiple inaccurate targets containing various information consistent with our prior knowledge about the true target are available. Rules 2-4 indicate: 1) Comparisons among small enough $LMP_{ts,ms}$ values generated by LAF-MPE for various methods can reasonably show not only the logical but also the true superiorities of corresponding methods compared with one another; 2) Comparisons among large enough $LMP_{ts,ms}$ values generated by LAF-MPE for various methods cannot confidently show the true but logical superiorities of these methods compared with one another. In summary, for the situation where accurate ground-truth labels (AGTLs) are not available while multiple inaccurate targets containing various information consistent with our prior knowledge about the true target are available, we have following practicability of LAF:

**Practicability 1**. LAF can be applied for evaluations with IAGTLs on a more difficult task, able to act like usual strategies for evaluations with AGTLs reasonably.

**Practicability 2**. LAF can be applied for evaluations with IAGTLs simply from the logical point of view on an easier task, unable to act like usual strategies for evaluations with AGTLs confidently.

## 6. Discussion

In this article, we proposed logical assessment formula for evaluations with inaccurate ground-truth labels (IAGTLs). Since LAF produces a series of logical assessment metrics of the predicted target compared with the underlying true target by referring to multiple inaccurate targets that contain various information consistent with our prior knowledge about the underlying true target, LAF achieves evaluations without the need of AGTLs. While some existing methods for evaluations with IAGTLs are proposed based on the assumption that some true targets probably exist in the inaccurate targets represented by the given IAGTLs, LAF does not require such fundamental assumption to achieve evaluations with IAGTLs. As a result, LAF offers a new addition to usual evaluations that require more or less AGTLs [1-5] as well as some existing methods for evaluations with IAGTLs [11–14].

Via logical reasoning under uncertainty, we also revealed the principles of LAF for evaluations with inaccurate ground-truth labels (IAGTLs). Different from the computational formalisms of reasoning under uncertainty established in existing articles [18–22] for computational intelligence, the logical reasoning under uncertainty, which is a series of logical processes with reasoning under uncertainty, conducted in this article more fundamentally / theoretically / cognitively aimed at revealing the principles of LAF (i.e., logical conclusions that can reflect the practicability of LAF) for evaluations with IAGTLs.

According to the principle revealed in section 4.1, the logical assessment metrics produced by LAF can reflect the logical rationality of the predicted target regarding to the underlying true target. With a series of logical assessment metrics produced by task-specific and method-specific LAF, the proposed LAF-MPE can achieve the method performance evaluation with IAGTLs. According to the principles revealed in section 4.2, LAF-MPE can reflect not only the method performance from the logical point of view but also under circumstances the true method performance regarding to the underlying true target.

The Practicability 1 of LAF summarized in Section 5 shows that LAF is able to act like usual strategies for evaluations with AGTLs reasonably on a more difficult task. According to the principles of LAF revealed in Section 4, this practicability of LAF indicates that it is suitable to achieve evaluations with IAGTLs for the situation where the underlying true targets are difficult to be precisely defined or even do not exist, since this situation usually contains relatively more difficult tasks which requires intensive labour for the label preparation in practice. However, in the meantime, the Practicability 2 of LAF summarized in Section 5 also shows that LAF is unable to act like usual strategies for evaluations with AGTLs confidently on an easier task. Similarly, according to the principles of LAF revealed in Section 4, this practicability of LAF also indicates that it is not suitable to achieve evaluations with IAGTLs for the situation where the true targets can easily be defined, since this situation usually contains relatively easier tasks which requires much less labour for the label preparation in practice.

Notably, though LAF requires multiple inaccurate targets extracted from IAGTLs to be carried out, it can be easily applied to achieve evaluations with IAGTLs. This is because the multiple inaccurate targets extracted from IAGTLs are merely required to contain various information consistent with our prior knowledge about the underlying true targets, which can be conveniently provided by experts when preparing the IAGTLs for a specific task. As a result, the preparation of IAGTLs for the application of LAF can be much less difficult and less labour-tensive.

## 7. Future Work

The first future work that should be done is to validate the revealed practicability of LAF for evaluations with IAGTLs to show its potentials for applications in practice. Since we are familiar with the field of medical histopathology whole slide image analysis, we have planned to apply LAF to tumour segmentation for breast cancer based on medical histopathology whole slide images for the validation of the revealed practicability of LAF for evaluations with IAGTLs.

The revealed practicability of LAF indicates that LAF is more suitable for evaluations with IAGTLs on more difficult tasks than on easier tasks. One natural question that can be raised is how can we judge a given task is more difficult or easer? In practice, we can qualitatively judge the difficulty of the given task based on some prior knowledge that has been accumulated for the given task, such as complaints of experts for how difficult it is to label the true targets for the given task. As qualitative judgement sometimes can be subjective, exploring solutions for quantitatively judging the difficulty of the given task is just a second work that should be done in the future.

In addition, the possible solutions to judging the difficulty of the given task can provide more evidences and clues to systematically investigate the limits for the application of LAF in practice, which is another work that is worth to be done in the future. We also look forward to possibly applying LAF to more other fields than simply the field of medical histopathology whole slide image analysis we are familiar with.

## Acknowledgement

The author, Yongquan Yang, especially thanks his Yang Family, Chengdu, China for providing him with the financial supports and mental encouragements to do this research.

## Reference

1. Chang HH, Zhuang AH, Valentino DJ, Chu WC (2009) Performance measure characterization for evaluating neuroimage segmentation algorithms. Neuroimage. https://doi.org/10.1016/j.neuroimage.2009.03.068

2. Taha AA, Hanbury A (2015) Metrics for evaluating 3D medical image segmentation: analysis, selection, and tool. BMC Med Imaging 15:29. https://doi.org/10.1186/s12880-015-0068-x

3. M H, M.N S (2015) A Review on Evaluation Metrics for Data Classification Evaluations. Int J Data Min Knowl Manag Process 5:01–11. https://doi.org/10.5121/ijdkp.2015.5201

4. Jung HJ, Lease M (2012) Evaluating Classifiers Without Expert Labels. https://doi.org/10.48550/arxiv.1212.0960

5. Deng W, Zheng L (2021) Are Labels Always Necessary for Classifier Accuracy Evaluation? In: Proceedings of the IEEE/CVF Conference on Computer Vision and Pattern Recognition (CVPR). pp 15069–15078

6. Warfield S, Dengler J, Zaers J, et al (1995) Automatic identification of gray matter structures from MRI to improve the segmentation of white matter lesions. J Image Guid Surg. https://doi.org/10.1002/(SICI)1522-712X(1995)1:6<326::AID-IGS4>3.0.CO;2-C

7. Kikinis R, Shenton ME, Gerig G, et al (1992) Routine quantitative analysis of brain and cerebrospinal fluid spaces with MR imaging. J Magn Reson Imaging. https://doi.org/10.1002/jmri.1880020603

8. Alonzo TA, Pepe MS (1999) Using a combination of reference tests to assess the accuracy of a new diagnostic test. Stat Med. https://doi.org/10.1002/(SICI)1097-0258(19991130)18:22<2987::AID-SIM205>3.0.CO;2-B

9. Beiden S V., Campbell G, Meier KL, Wagner RF (2000) The problem of ROC analysis without truth: the EM algorithm and the information matrix. In:

Krupinski EA (ed) Medical Imaging 2000: Image Perception and Performance. pp 126–134

10. Korevaar DA, Toubiana J, Chalumeau M, et al (2021) Evaluating tests for diagnosing COVID-19 in the absence of a reliable reference standard: pitfalls and potential solutions. J Clin Epidemiol. https://doi.org/10.1016/j.jclinepi.2021.07.021

11. Warfield SK, Zou KH, Wells WM (2004) Simultaneous truth and performance level estimation (STAPLE): An algorithm for the validation of image segmentation. IEEE Trans Med Imaging. https://doi.org/10.1109/TMI.2004.828354

12. Martin-Fernandez M, Bouix S, Ungar L, et al (2005) Two Methods for Validating Brain Tissue Classifiers. In: Lecture Notes in Computer Science (including subseries Lecture Notes in Artificial Intelligence and Lecture Notes in Bioinformatics). pp 515–522

13. Bouix S, Martin-Fernandez M, Ungar L, et al (2007) On evaluating brain tissue classifiers without a ground truth. Neuroimage. https://doi.org/10.1016/j.neuroimage.2007.04.031

14. Joyce RJ, Raff E, Nicholas C (2021) A Framework for Cluster and Classifier Evaluation in the Absence of Reference Labels. In: Proceedings of the 14th ACM Workshop on Artificial Intelligence and Security. ACM, New York, NY, USA, pp 73–84

15. Yang Y, Yang Y, Yuan Y, et al (2020) Detecting helicobacter pylori in whole slide images via weakly supervised multi-task learning. Multimed Tools Appl 79:26787–26815. https://doi.org/10.1007/s11042-020-09185-x

16. Yang Y, Yang Y, Chen J, et al (2020) Handling Noisy Labels via One-Step Abductive Multi-Target Learning and Its Application to Helicobacter Pylori Segmentation

17. Zhou ZH (2019) Abductive learning: towards bridging machine learning and logical reasoning. Sci China Inf Sci. https://doi.org/10.1007/s11432-018-9801-4

18. Pearl J (1990) Reasoning Under Uncertainty. Annu Rev Comput Sci 4:37–72. https://doi.org/10.1146/annurev.cs.04.060190.000345

19. Krause P, Ambler S, Elvang-Goransson M, Fox J (1995) A LOGIC OF ARGUMENTATION FOR REASONING UNDER UNCERTAINTY. Comput Intell 11:113–131. https://doi.org/10.1111/j.1467-8640.1995.tb00025.x

20. Parsons S (2001) Qualitative Methods for Reasoning under Uncertainty. The MIT Press

21. Dubois D, Prade H, Schockaert S (2017) Generalized possibilistic logic: Foundations and applications to qualitative reasoning about uncertainty. Artif Intell 252:139–174. https://doi.org/10.1016/j.artint.2017.08.001

22. Ristic B, Gilliam C, Byrne M (2021) Performance assessment of a system for reasoning under uncertainty. Inf Fusion 71:11–16. https://doi.org/10.1016/j.inffus.2021.01.006

23. Müller H, Holzinger A (2021) Kandinsky Patterns. Artif Intell 300:103546. https://doi.org/10.1016/j.artint.2021.103546

24. Dempster AP, Laird NM, Rubin DB (1977) Maximum Likelihood from Incomplete Data Via the EM Algorithm . J R Stat Soc Ser B. https://doi.org/10.1111/j.2517-6161.1977.tb01600.x

25. Warfield SK, Zou KH, Wells WM (2002) Validation of image segmentation and expert quality with an expectation-maximization algorithm. In: Lecture Notes in Computer Science (including subseries Lecture Notes in Artificial Intelligence and Lecture Notes in Bioinformatics)

26. Beynon M, Curry B, Morgan P (2000) The Dempster–Shafer theory of evidence: an alternative approach to multicriteria decision modelling. Omega 28:37–50. https://doi.org/10.1016/S0305-0483(99)00033-X

# Supplementary

## Preliminary of Logical Reasoning

We introduce some propositional connectives and rules for proof of propositional logical reasoning, which are respectively shown as Table 1 and Table 2, for the logical reasonings conducted in this paper.

Table 1. Propositional connectives

| Connective | Meaning |
|---|---|
| $\wedge$ | conjunction |
| $\rightarrow$ | implication |

Table 2. Rules for proof of propositional logical reasoning, $\vdash$ denotes 'bring out'

| Rule | Meaning |
|---|---|
| $\wedge -$ | reductive law of conjunction: $A \wedge B$, $\vdash A$ or B. |
| $\wedge +$ | additional law of conjunction: $A, B$, $\vdash A \wedge B$. |
| MP | modus ponens: $A \rightarrow B$, $A$, $\vdash B$. |
| HS | hypothetical syllogism: $A \rightarrow B$, $B \rightarrow C$, $\vdash A \rightarrow C$. |

## Proof of Theorem 1

**Theorem 1**. *If the input $t$ of LAF is the prediction for the underlying true target which is indefinable, and the input $\tilde{t}$ of LAF has multiple inaccurate targets containing various information consistent with our prior knowledge about the underlying true target, then the output of LAF ($LAM$) can reflect the logical rationality of the input $t$ of LAF compared with the underlying true target.*

**Proof-T1**. Firstly, with LAF, we have following derived preconditions for Theorem 1:

1. If the input $\tilde{t}$ of LAF has multiple inaccurate targets containing various information consistent with our prior knowledge about the underlying true target, then logical facts ($LF$) can be narrated from the input $\tilde{t}$ via the logical fact narration component of LAF.
2. If logical facts ($LF$) can be narrated from the input $\tilde{t}$ via the logical fact narration component of LAF, and the input $\tilde{t}$ has multiple inaccurate targets containing various information consistent with our prior knowledge about the underlying true target, then the narrated $LF$ can also contain some information consistent with our prior knowledge about the underlying true target.
3. If the narrated $LF$ can also contain some information consistent with our prior knowledge about the underlying true target, then the characteristics of the narrated $LF$ are consistent with some characteristics of the underlying true target.
4. If the input $t$ of LAF is the prediction for the underlying true target which is indefinable, and $LF$ can be narrated from the input $\tilde{t}$ via the logical fact narration component of LAF, then logical consistencies ($LC$) between the input $t$ and the narrated $LF$ can be estimated via the logical consistency estimation component of LAF.

5. If $LC$ between the input $t$ and the narrated $LF$ can be estimated via the logical consistency estimation component of LAF, then the estimated $LC$ can describe the logical consistencies between the characteristics of the input $t$ and the characteristics of the narrated $LF$.
6. If the estimated $LC$ can describe the logical consistencies between the characteristics of the input $t$ and the characteristics of the narrated $LF$, and the characteristics of the narrated $LF$ are consistent with some characteristics of the underlying true target, then the estimated $LC$ can describe the logical consistencies between the characteristics of the input $t$ and some characteristics of the underlying true target.
7. If the estimated $LC$ can describe the logical consistencies between the characteristics of the given $t$ and some characteristics of the underlying true target, then the estimated $LC$ can reflect the logical rationality of the input $t$ compared with the underlying true target.
8. If $LC$ between the input $t$ and the narrated $LF$ can be estimated via the logical consistency estimation component of LAF, then the logical assessment metrics ($LAM$) of the input $t$ can be built based on the estimated $LC$ via the logical assessment metric build component of LAF.
9. If $LAM$ of the input $t$ can be built based on the estimated $LC$ via the logical assessment metric build component of LAF, then the built $LAM$ is derived from the estimated $LC$.
10. If the built $LAM$ is derived from the estimated $LC$, and the estimated $LC$ can reflect the logical rationality of the given $t$ compared with the underlying true target, then the build $LAM$ can also reflect the logical rationality of the input $t$ compared with the underlying true target.
11. If the build $LAM$ can also reflect the logical rationality of the input $t$ compared with the underlying true target, then the output of LAF ($LAM$) can reflect the logical rationality of the input $t$ of LAF compared with the underlying true target.

Secondly, we give the propositional symbols for the above preconditions 1-11 for Theorem 1, which are shown in Table 3.

Table 3. Propositional symbols of preconditions for Theorem 1

| Symbol | Meaning |
|---|---|
| $a$ | the input $\tilde{t}$ of LAF has multiple inaccurate targets containing various information consistent with our prior knowledge about the underlying true target |
| $b$ | logical facts ($LF$) can be narrated from the input $\tilde{t}$ via the logical fact narration component of LAF |
| $c$ | the narrated $LF$can also contain some information consistent with our prior knowledge about the underlying true target |
| $d$ | the characteristics of the narrated $LF$ are consistent with some characteristics of the underlying true target |

| $e$ | the input $t$ of LAF is the prediction for the underlying true target which is indefinable |
|---|---|
| $f$ | logical consistencies ($LC$) between the input $t$ and the narrated $LF$ can be estimated via the logical consistency estimation component of LAF |
| $g$ | the estimated $LC$ can describe the logical consistencies between the characteristics of the input $t$ and the characteristics of the narrated $LF$ |
| $h$ | the estimated $LC$ can describe the logical consistencies between the characteristics of the input $t$ and some characteristics of the underlying true target |
| $i$ | the estimated $LC$ can reflect the logical rationality of the input $t$ compared with the underlying true target |
| $j$ | the logical assessment metrics ($LAM$) of the input $t$ can be built based on the estimated $LC$ via the logical assessment metric build component of LAF |
| $k$ | the built $LAM$ is derived from the estimated $LC$ |
| $l$ | the build $LAM$ can also reflect the logical rationality of the given $t$ compared with the underlying true target |
| $m$ | the output of LAF ($LAM$) can reflect the logical rationality of the input $t$ of LAF compared with the underlying true target |

Thirdly, referring to Table 3, we signify the propositional formalizations of the preconditions 1-11 for Theorem 1 and Theorem 1 via the propositional connectives listed in Table 1 as follows.

1) $a \rightarrow b$ Precondition
2) $(a \wedge b) \rightarrow c$ Precondition
3) $c \rightarrow d$ Precondition
4) $(e \wedge b) \rightarrow f$ Precondition
5) $f \rightarrow g$ Precondition
6) $(g \wedge d) \rightarrow h$ Precondition
7) $h \rightarrow i$ Precondition
8) $f \rightarrow j$ Precondition
9) $j \rightarrow k$ Precondition
10) $(k \wedge i) \rightarrow l$ Precondition
11) $l \rightarrow m$ Precondition

$(e \wedge a) \rightarrow m$ Theorem 1

Fourthly, we show the validity of Theorem 1 via the rules for proof of propositional logical reasoning listed in Table 2 as follows.

$\therefore \boldsymbol{(e \wedge a) \rightarrow m}$

12) $e \wedge a$ Hypothesis
13) $e$ 12); $\wedge -$
14) $a$ 12); $\wedge -$
15) $b$ 1),10); MP
16) $a \wedge b$ 14),15); $\wedge +$
17) $(a \wedge b) \rightarrow d$ 2),3); HS
18) $d$ 17),16); MP
19) $e \wedge b$ 13),15); $\wedge +$
20) $(e \wedge b) \rightarrow g$ 4),5); HS
21) $g$ 20),19); MP

22) $g \wedge d$ — 21),18); $\wedge +$

23) $(g \wedge d) \rightarrow i$ — 6),7); HS

24) $i$ — 23),22); MP

25) $(e \wedge b) \rightarrow k$ — 4),8),9); HS

26) $k$ — 25),19); MP

27) $k \wedge i$ — 26),24); $\wedge +$

28) $l$ — 10),27); MP

29) $m$ — 11),28); MP

30) $(e \wedge a) \rightarrow m$ — 12)-29); Conditional Proof

Since the hypothesis $e \wedge a$ of the 12) step indeed can be fulfilled in practice [1,2], Theorem 1 is proved to be valid.

## Proof of Theorem 2

**Theorem 2**. *If the input $LAM_{ts,ms}$ of LAF-MPE is a series of logical assessment metrics produced by task-specific and method-specific LAF, then the output $LMP_{ts,ms}$ of LAF-MPE can reflect the method performance from the logical point of view.*

**Proof-T2**. Firstly, with the proposed LAF-MPE, we have following derived preconditions:

1. If the input $LAM_{ts,ms}$ of LAF-MPE is a series of logical assessment metrics produced by task-specific and method-specific LAF, then the input $LAM_{ts,ms}$ can reflect the logical rationality of the given specific method for the given specific task according to Theorem 1, and the evaluation of the given specific method for the given specific task ($LMP_{ts,ms}$) can be estimated form the input $LAM_{ts,ms}$ via the component of LAF-MPE.
2. If the evaluation of the given specific method for the given specific task ($LMP_{ts,ms}$) can be estimated form the input $LAM_{ts,ms}$ via the component of LAF-MPE, then the estimated $LMP_{ts,ms}$ can in a way reflect the performance of the given specific method for the given specific task, the evaluated $LMP_{ts,ms}$ is derived from $LAM_{ts,ms}$, and the evaluated $LMP_{ts,ms}$ is the output of LAF-MPE.
3. If the estimated $LMP_{ts,ms}$ can in a way reflect the performance of the given specific method for the given specific task, the input $LAM_{ts,ms}$ can reflect the logical rationality evaluations of the given specific method for the given specific task according to Theorem 1, and the estimated $LMP_{ts,ms}$ is derived from $LAM_{ts,ms}$, then the estimated $LMP_{ts,ms}$ can reflect the performance of the given specific method for the given specific task from the logical point of view.
4. If the estimated $LMP_{ts,ms}$ can reflect the performance of the given specific method for the given specific task from the logical point of view and the estimated $LMP_{ts,ms}$ is the output of LAF-MPE, then the output $LMP_{ts,ms}$ of LAF-MPE can reflect the method performance from the logical point of view.

Secondly, we give the propositional symbols for the above preconditions 1-4 for Theorem 2, which are shown in Table 4.

Table 4. Propositional symbols of preconditions for Theorem 2

| Symbol | Meaning |
|---|---|
| $o$ | the input $LAM_{ts,ms}$ of LAF-MPE is a series of logical assessment metrics produced by task-specific and method-specific LAF |
| $p$ | the input $LAM_{ts,ms}$ can reflect the logical rationality of the given specific method for the given specific task according to Theorem 1 |
| $q$ | the evaluation of the given specific method for the given specific task ($LMP_{ts,ms}$) can be estimated form the input $LAM_{ts,ms}$ via the component of LAF-MPE |
| $r$ | the estimated $LMP_{ts,ms}$ can in a way reflect the performance of the given specific method for the given specific task |
| $s$ | the estimated $LMP_{ts,ms}$ is derived from $LAM_{ts,ms}$ |
| $t$ | the estimated $LMP_{ts,ms}$ is the output of LAF-MPE |
| $u$ | the estimated $LMP_{ts,ms}$ of LAF-MPE can reflect the performance of the given specific method for the given specific task from logical point of view |
| $v$ | the output $LMP_{ts,ms}$ of LAF-MPE can reflect the method performance from logical point of view |

Thirdly, referring to Table 4, we signify the propositional formalizations of the preconditions 1-4 for Theorem 2 and Theorem 2 via the propositional connectives listed in Table 1 as follows.

| | |
|---|---|
| 1) $o \rightarrow (p \wedge q)$ | Precondition |
| 2) $q \rightarrow (r \wedge s \wedge t)$ | Precondition |
| 3) $(r \wedge p \wedge s) \rightarrow u$ | Precondition |
| 4) $(u \wedge t) \rightarrow v$ | Precondition |
| $o \rightarrow v$ | Theorem 2 |

Fourthly, we show the validity of Theorem 2 via the rules for proof of propositional logical reasoning listed in Table 2 as follows.

$\therefore \boldsymbol{o \rightarrow v}$

| | |
|---|---|
| 5) $o$ | Hypothesis |
| 6) $p \wedge q$ | 1),5); MP |
| 7) $p$ | 6); $\wedge -$ |
| 8) $q$ | 6); $\wedge -$ |
| 9) $r \wedge s \wedge t$ | 2),8); MP |
| 10) $r$ | 9); $\wedge -$ |
| 11) $s$ | 9); $\wedge -$ |
| 12) $t$ | 9); $\wedge -$ |
| 13) $r \wedge p \wedge s$ | 10),7),11); $\wedge +$ |
| 14) $u$ | 3),13); MP |
| 15) $u \wedge t$ | 14),12); $\wedge +$ |
| 16) $v$ | 4),15); MP |
| 17) $o \rightarrow v$ | 5)-16); Conditional Proof |

Since the hypothesis $o$ of the 5) step can be fulfilled by LAF, Theorem 2 is proved to be valid.

## Preliminary for proofs of Theorem 3 and Theorem 4

Before presenting the proofs of Theorem 3 and Theorem 4, we give two postulates for evaluating the overall performance of a given specific method for a given specific task. We give Postulate 1, as the true performance of the given specific method for the given specific task can be evaluated from various perspectives. We give Postulate 2, as evaluations from various perspectives can usually be carried out with the same prediction of the given specific method for the given specific task.

**Postulate 1**. The true performance of a given specific method for a given specific task can be evaluated from logical perspective and other than logical perspective.

**Postulate 2**. Correlation exists between the evaluation from logical perspective and the evaluation from other than logical perspective.

Based on Postulate 1 and Postulate 2, we introduce following three lemmas:

**Lemma 1**. If $LPE$ denotes the evaluation from logical perspective, then $c \cup LPE$ (complementary set of $LPE$) denotes the evaluation from other than logical perspective.

**Lemma 2**. If $Val(LPE)$ denotes the value of evaluation form logical perspective, and $Val(c \cup LPE)$ denotes the value of evaluation from other than logical perspective, then the value of the true performance can be formalized as $Val(LPE) \times Val(c \cup LPE)$.

**Lemma 3**. If the output $LMP_{ts,ms}$of LAF-MPE can reflect the method performance from logical point of view, then the value of the true method performance can be formalized as $Val(LMP_{ts,ms}) \times Val(c \cup LMP_{ts,ms})$.

Lemma 1, Lemma 2, and Lemma 3 are respectively proved by following proofs (Proof-L1, Proof-L2, Proof-L3).

**Proof-L1**. Let $A$ denote the true performance evaluation, $B$ denote the evaluation from logical perspective, and $C$ denote the evaluation form other than logical perspective. From Postulate 1, we have $A = B \cup C$. Referring to $A = B \cup C$, we can infer that $C$ is the complementary set of $B$, i.e., $C = c \cup B$. As a result, the proposition that if $B$ denote the evaluation from logical perspective then $c \cup B$ denotes the evaluation from other than logical perspective is true. Let $LPE$ replace $B$, then the proposition is true that if $LPE$ denotes the evaluation from logical perspective then $c \cup LPE$ denotes the evaluation from other than logical perspective. Thus, Lemma 1 is true.

**Proof-L2**. Let $Val(A)$ denote the value of the true performance evaluation ($A$), $Val(B)$ denote the value of evaluation from logical perspective ($B$), and $Val(C)$ denote the value of evaluation from other than logical perspective ($C$). Based on Postulate 1, we have $A = B \cup C$. Based on Postulate 2, we have $B \cap C \neq \emptyset$. Referring to $A = B \cup C$, we can formalize the equation for the value of the true performance by $Val(A) = f\big(Val(B), Val(C)\big)$, where $f$ is the function with $Val(B)$ and $Val(C)$ as input variables. Referring to $B \cap C \neq \emptyset$, we can assume that the correlation between $Val(B)$ and $Val(C)$ is associated with $Val(A)$. Referring to the perspective that multiplicative model assumes that correlations among the input variables are associated with the outcome indicator [3], the formalization of $f$ can be a multiplicative model, i.e., $f$ can be formalized as $\prod_{i=1}^{n} i$. As a result, we can have $Val(B) \times Val(C)$ to formalize the value of the true performance. Let $LE$ replace $B$ and $c \cup LE$ replace $C$, the proposition is true that if $Val(LPE)$ denotes the value of evaluation form logical perspective, and $Val(c \cup LPE)$ denotes the value of evaluation from other than logical perspective, then the value of the true performance can be formalized as $Val(LPE) \times Val(c \cup LPE)$. Thus Lemma 2 is true.

**Proof-L3**. Firstly, with the proposed LAF-MPE, Lemma 1 and Lemma 2, we have following derived common preconditions for Lemma 3:

1. If the output $LMP_{ts,ms}$of LAF-MPE can reflect the method performance from logical point of view, then $LMP_{ts,ms}$ denotes the method evaluation from logical perspective.
2. If $LMP_{ts,ms}$ denotes the method evaluation from logical perspective, then the value of $LMP_{ts,ms}$ ($Val(LMP_{ts,ms})$) denotes the value of the method evaluation form logical perspective.
3. If $LMP_{ts,ms}$ denotes the method evaluation from logical perspective, then $c \cup LMP_{ts,ms}$ (complementary set of $LMP_{ts,ms}$) denotes the method evaluation from other than logical perspective. (Lemma 1)
4. If $c \cup LMP_{ts,ms}$ denotes the method evaluation from other than logical perspective, then the value of $c \cup LMP_{ts,ms}$ ($Val(c \cup LMP_{ts,ms})$) denotes the value of method evaluation from other than logical perspective.
5. If $Val(LMP_{ts,ms})$ denotes the value of the method evaluation form logical perspective and $Val(c \cup LMP_{ts,ms})$ denotes the value of the method evaluation from other than logical perspective, then the value of the true method performance can be formalized as $Val(LMP_{ts,ms}) \times Val(c \cup LMP_{ts,ms})$. (Lemma 2)

Secondly, we give the propositional symbols for the above preconditions 1-5 for Lemma3, which are shown in Table 5.

Table 5. Propositional symbols of preconditions for Lemma 3

| Symbol | Meaning |
|---|---|
| $w$ | the output $LMP_{ts,ms}$of LAF-MPE can reflect the method performance from logical point of view |
| $x$ | $LMP_{ts,ms}$ denotes the method evaluation from logical perspective |
| $y$ | the value of $LMP_{ts,ms}$ ($Val(LMP_{ts,ms})$) denotes the value of the method evaluation form logical perspective |
| $z$ | $c \cup LMP_{ts,ms}$ (complementary set of $LMP_{ts,ms}$) denotes the method evaluation from other than logical perspective |
| $a$ | the value of $c \cup LMP_{ts,ms}$ ($Val(c \cup LMP_{ts,ms})$) denotes the value of the method evaluation from other than logical perspective |
| $b$ | the value of the true method performance can be formalized as $Val(LMP_{ts,ms}) \times Val(c \cup LMP_{ts,ms})$ |

Thirdly, referring to Table 5, we signify the propositional formalizations of the preconditions 1-5 for Lemma 3 and Lemma 3 via the propositional connectives listed in Table 1 as follows.

1) $w \to x$ Precondition

2) $x \rightarrow y$ Precondition

3) $x \rightarrow z$ Precondition

4) $z \rightarrow a$ Precondition

5) $(y \wedge a) \rightarrow b$ Precondition

$w \rightarrow b$ Lemma 3

Fourthly, we show the validity of Lemma 3 via the rules for proof of propositional logical reasoning listed in Table 2 as follows.

$\therefore \boldsymbol{w \rightarrow b}$

6) $w$ Hypothesis

7) $x$ 1),6); MP

8) $y$ 2),7); MP

9) $x \rightarrow a$ 3),4); HS

10) $a$ 9),7); MP

11) $y \wedge a$ 8),10); $\wedge +$

12) $b$ 5),11); MP

13) $w \rightarrow b$ 6)-12); Conditional Proof

Since the hypothesis $w$ of the 6) step can be fulfilled by Theorem 2, Lemma 3 is proved to be valid.

## Proof of Theorem 3

**Theorem 3**. *If the output $LMP_{ts,ms}$ of LAF-MPE can reflect the method performance from the logical point of view and the value of $LMP_{ts,ms}$ is small enough, then the output $LMP_{ts,ms}$ of LAF-MPE can reflect the true method performance reasonably.*

**Proof-T3**. Firstly, on the basis of Lemma 3, we have following preconditions for Theorem 3.

1. If the output $LMP_{ts,ms}$ of LAF-MPE can reflect the method performance from the logical point of view, then the value of the true method performance can be formalized as $Val(LMP_{ts,ms}) \times Val(c \cup LMP_{ts,ms})$. (Lemma 3)
2. If the value of the true method performance can be formalized as $Val(LMP_{ts,ms}) \times Val(c \cup LMP_{ts,ms})$ and the value of $LMP_{ts,ms}$ is small enough, then small enough value of $LMP_{ts,ms}$ can bring out relatively small value of the true method performance.
3. If small enough value of $LMP_{ts,ms}$ can bring out relatively small value of the true method performance, then the output $LMP_{ts,ms}$ of LAF-MPE can reflect the true method performance reasonably.

Secondly, we give the propositional symbols for the above preconditions 1-3 for Theorem 3, which are shown in Table 6.

Table 6. Propositional symbols of preconditions for Theorem 3

| Symbol | Meaning |
|---|---|
| $c$ | the output $LMP_{ts,ms}$ of LAF-MPE can reflect the method performance from the logical point of view |
| $d$ | the value of the true method performance can be formalized as $Val(LMP_{ts,ms}) \times Val(c \cup LMP_{ts,ms})$ |
| $e$ | the value of $LMP_{ts,ms}$ is small enough |
| $f$ | small enough value of $LMP_{ts,ms}$ can bring out relatively small value of the true method performance |
| $g$ | the output $LMP_{ts,ms}$ of LAF-MPE can reflect the true method performance reasonably |

Thirdly, referring to Table 6, we signify the propositional formalizations of the preconditions 1-3 for Theorem 3 and Theorem 3 via the propositional connectives listed in Table 1 as follows.

| | |
|---|---|
| 1) $c \to d$ | Precondition |
| 2) $(d \land e) \to f$ | Precondition |
| 3) $f \to g$ | Precondition |
| $(c \land e) \to g$ | Theorem 3 |

Fourthly, we show the validity of Theorem 3 via the rules for proof of propositional logical reasoning listed in Table 2 as follows.

$\therefore (\boldsymbol{c} \land \boldsymbol{e}) \to \boldsymbol{g}$

| | |
|---|---|
| 4) $c \land e$ | Hypothesis |
| 5) $c$ | 4); $\land -$ |
| 6) $e$ | 4); $\land -$ |
| 7) $d$ | 1),5); MP |
| 8) $d \land e$ | 7),6); $\land +$ |
| 9) $(d \land e) \to g$ | 2),3); HS |
| 10) $g$ | 9),8); MP |
| 11) $c \land e \to g$ | 4)-10); Conditional Proof |

Since the hypothesis $c$ of the 4) step can be fulfilled by Theorem 2 and the hypothesis $e$ of the 4) step can be fulfilled by thresholding the value of the output of LAF-MPE ($LMP_{ts,ms}$), Theorem 3 is proved to be valid.

## Proof of Theorem 4

**Theorem 4**. *If the output $LMP_{ts,ms}$ of LAF-MPE can reflect the method performance from the logical point of view and the value of $LMP_{ts,ms}$ is large enough, then the output $LMP_{ts,ms}$ of LAF-MPE cannot reflect the true method performance confidently.*

**Proof-T4**. Firstly, on the basis of Lemma 3, we have following preconditions for Theorem 4.

1. If the output $LMP_{ts,ms}$ of LAF-MPE can reflect the method performance from the logical point of view, then the value of the true method performance can be formalized as $Val(LMP_{ts,ms}) \times Val(c \cup LMP_{ts,ms})$. (Lemma 3)
2. If the value of the true method performance can be formalized as $Val(LMP_{ts,ms}) \times Val(c \cup LMP_{ts,ms})$, and the value of $LMP_{ts,ms}$ is large enough, then large enough value of $LMP_{ts,ms}$ cannot bring out large value of the true method performance confidently.

3. If large enough value of $LMP_{ts,ms}$ cannot bring out large value of the true method performance confidently, then the output $LMP_{ts,ms}$ of LAF-MPE cannot reflect the true method performance confidently.

Secondly, we give the propositional symbols for the above preconditions 1-3 for Theorem 4, which are shown in Table 7.

Table 7. Propositional symbols of preconditions for Theorem 4

| Symbol | Meaning |
|---|---|
| $h$ | the output $LMP_{ts,ms}$ of LAF-MPE can reflect the method performance from the logical point of view |
| $i$ | the value of the true method performance can be formalized as $Val(LMP_{ts,ms}) \times Val(c \cup LMP_{ts,ms})$ |
| $j$ | the value of $LMP_{ts,ms}$ is large enough |
| $k$ | large enough value of $LMP_{ts,ms}$ cannot bring out large value of the true method performance confidently |
| $l$ | the output $LMP_{ts,ms}$ of LAF-MPE cannot reflect the true method performance confidently |

Thirdly, referring to Table 6, we signify the propositional formalizations of the preconditions 1-3 for Theorem 3 and Theorem 3 via the propositional connectives listed in Table 1 as follows.

1) $h \to i$ Precondition
2) $(i \wedge j) \to k$ Precondition
3) $k \to l$ Precondition
$(h \wedge j) \to l$ Theorem 4

Fourthly, we show the validity of Theorem 4 via the rules for proof of propositional logical reasoning listed in Table 2 as follows.

$\therefore (\boldsymbol{h} \wedge \boldsymbol{j}) \to \boldsymbol{l}$

4) $h \wedge j$ Hypothesis
5) $h$ 4); $\wedge -$
6) $j$ 4); $\wedge -$
7) $i$ 1),5); MP
8) $i \wedge j$ 7),6); $\wedge +$
9) $(i \wedge j) \to l$ 2),3); HS
10) $l$ 9),8); MP
11) $(h \wedge j) \to l$ 4)-10); Conditional Proof

Since the hypothesis $h$ of the 4) step can be fulfilled by Theorem 2 and the hypothesis $j$ of the 4) step can be fulfilled by thresholding the value of the output of LAF-MPE ($LMP_{ts,ms}$), Theorem 4 is proved to be valid.

## Reference

[1] Y. Yang, Y. Yang, Y. Yuan, J. Zheng, Z. Zhongxi, Detecting helicobacter pylori in whole slide images via weakly supervised multi-task learning, Multimed. Tools Appl. 79 (2020) 26787–26815. https://doi.org/10.1007/s11042-020-09185-x.

[2] Y. Yang, Y. Yang, J. Chen, J. Zheng, Z. Zheng, Handling Noisy Labels via One-Step Abductive Multi-Target Learning and Its Application to Helicobacter Pylori Segmentation, (2020). http://arxiv.org/abs/2011.14956.

[3] T.E. Becker, J.A. Cote, Additive and Multiplicative Method Effects in Applied Psychological Research: An Empirical Assessment of Three Models, J. Manage. (1994). https://doi.org/10.1177/014920639402000306.